\pdfoutput=1

\documentclass[11pt]{article}

 \usepackage[preprint]{acl}
\usepackage{times}
\usepackage{latexsym}
\usepackage{booktabs}

\usepackage[T1]{fontenc}

\usepackage[utf8]{inputenc}

\usepackage{microtype}

\usepackage{inconsolata}

\usepackage{graphicx}
\usepackage{amsmath}
\usepackage{multirow} 
 \usepackage{algorithm}
\usepackage{algpseudocode}
\usepackage{dialogue}

\usepackage{textcomp}
\usepackage{mathtools}
\usepackage{arydshln}
\usepackage{mathrsfs}
\usepackage{multirow}
\usepackage{subcaption}
\usepackage{xcolor}
\usepackage{soul}
\usepackage{tcolorbox}
\usepackage{caption}
\usepackage[inline]{enumitem}

\title{“Any Other Thoughts, Hedgehog?” Linking Deliberation Chains in Collaborative Dialogues}

\author{Abhijnan Nath, Videep Venkatesha, Mariah Bradford, Avyakta Chelle, \\ {\bf Austin Youngren, Carlos Mabrey, Nathaniel Blanchard, Nikhil Krishnaswamy} \\ Situated Grounding and Natural Language (SIGNAL) Lab, \\ Department of Computer Science, Colorado State University, Fort Collins, CO USA \\{\tt firstname.lastname@colostate.edu}}

\begin{document}
\maketitle
\begin{abstract}

Question-asking in collaborative dialogue has long been established as key to knowledge construction, both in internal and collaborative problem solving. In this work, we examine \textit{probing questions} in collaborative dialogues: questions that explicitly elicit responses from the speaker's interlocutors. Specifically, we focus on modeling the causal relations that lead directly from utterances earlier in the dialogue to the emergence of the probing question. We model these relations using a novel graph-based framework of {\it deliberation chains}, and reframe the problem of constructing such chains as a coreference-style clustering problem. Our framework jointly models probing and causal utterances and the links between them, and we evaluate on two challenging collaborative task datasets: the Weights Task and DeliData. Our results demonstrate the effectiveness of our theoretically-grounded approach compared to both baselines and stronger coreference approaches, and establish a standard of performance in this novel task.

\end{abstract}

\section{Introduction}
\label{sec:intro}
\vspace*{-2mm}

Recent breakthroughs in generative AI have raised the possibility of systems that follow and interact with multiparty dialogue. Inherent in group dialogues are utterance sequences that deliberate on the same information. Modeling these is particularly challenging; while such utterances have a linear order and overlapping information, they may be distantly separated in time and the same information may be expressed very differently.

In this paper, we construct \textbf{deliberation chains} in dialogue: turn sequences that surface pieces of evidence or questions under discussion that culminate in a ``probing utterance,'' or explicit elicitation of input that does not introduce new information. We model deliberation chains as {\it probing interventions} that are preceded somewhere in the discourse by a number of {\it causal interventions}, each of which contribute directly to the eventual emergence of the utterance that serves as the probing intervention. Without the causal counterpart(s), the probe would not arise in the discourse (at least not in that specific form or at that specific time). 

Both probe and cause are linked to effective group performance \cite{karadzhov2023delidata}. Tracking them requires an evolving understanding of collaborative dynamics and enables disagreement detection, prompting for deeper insights, or analysis of deliberation's influence on individual learning/understanding~\cite{hunter2018formal,atwell2024combining,khebour2024common}.   

Our novel approach  takes inspiration from discourse coherence theory 
and joint modeling frameworks traditionally applied to coreference resolution. The ability to link probing interventions to their preceding causes in the dialogue is a critical prerequisite for AI systems to support deliberative/collaborative reasoning, which is of interest in domains like education and workforce development.
Our novel contributions are:
  \begin{itemize}
    \vspace*{-2mm}
    \item A novel task of automatically constructing ``deliberation chains'' of probing questions in a dialogue and with their causal utterances;
    \vspace*{-2mm}
    \item A formal graphical framework for deliberation chains 
    derived from formal semantics of situated conversation~\cite{hunter2018formal};
    \vspace*{-2mm}
    \item A unique adaptation of methods from coreference resolution to this new task;
    \vspace*{-2mm}
    \item Baseline evaluation on two challenging collaborative dialogue datasets---DeliData and the Weights Task Dataset---and a novel method of jointly modeling probing and causal interventions and the links between them.
    
 
 \end{itemize}

Our code may be found at: \url{https://github.com/csu-signal/ProbingDelibration}

\section{Related Work}
\vspace*{-2mm}

\paragraph{Collaborative Dynamics}
\citet{andrews-todd_exploring_2020}, \citet{oecd_pisa_2017}, and \citet{sun_towards_2020} have all identified the need for teams to construct shared knowledge to function, often through asking questions. \citet{hesse_framework_2015} also points out that collaboration requires teammates to initiate interaction. Further, \citet{fusaroli_measures_2017} identified conversational repair as a necessary mechanism in forming common ground for a group. \citet{graesser_advancing_2018} describes the need for team members to externalize their knowledge. \citet{karadzhov2022makes} explores how deliberation may lead to team members changing their minds, which is critical for building group common ground \cite{stalnaker1978assertion}.


\paragraph{Joint Modeling in Coreference Resolution}
In the well-studied problem of coreference resolution, many works~\cite{lee-etal-2017-end,zhang2018neural, cattan2021cross, yu2022pairwise} have proposed various joint modeling frameworks and cross-encoding architectures that optimize on coreference link assignments and building mention clusters, including ideas to make such methods more scalable~\cite{ahmed20232, held-etal-2021-focus} and generalizable~\cite{bugert2021generalizing}. In contrast to such methods that often operate on a "span"-level and require exhaustive cross-computations~\cite{thirukovalluru2021scaling}, we generate deliberation chains using utterances as distinct discourse units.


\paragraph{Free-Text Rationales}
With the advent of instruction-tuned generative LLMs like InstructGPT~\cite{ouyang2022training}, 
recent works \cite{ahmed-etal-2024-linear-cross,wang2024reasoning,radhakrishnan2023question,zhao2023abductive} have leveraged their Chain-of-Thought (COT)-style reasoning capacities for various NLP tasks like argument extraction, question-answering as well as coreference annotations, often guiding the LLM's reasoning process using Free-Text Rationales (FTRs) that explicate reasoning steps toward a decision \cite{wiegreffe2021measuring, west-etal-2022-symbolic, nath2024okay}. Our work uses such FTRs to guide the automatic annotation of interventions in collaborative task datasets.


\section{Problem Formulation}
\label{sec:problem_formulation}
\vspace*{-2mm}


Segmented Discourse Representation Theory (SDRT) posits that interpreting an utterance involves supplementing its semantics with pragmatic content based on the demands of {\it discourse coherence}~\cite{asher2003logics}. 
The relation between utterances and prior content required for a full interpretation gives rise to structures which in collaborative dialogues represent the evolution of information that propels such dialogues towards task-completion~\cite{karadzhov2023delidata}. Let us define the relevant structures below:



\textbf{Definition 1.} Based on \citet{hunter2018formal}, let \( \mathcal{G} = (\mathcal{V}, \mathcal{E}_1, \mathcal{E}_2, \lambda) \) be a {\bf deliberation graph} in a collaborative dialogue, that in turn comprises sets of individual deliberation chains. \( \mathcal{G} \) is characterized as a weakly-connected, weighted, acyclic graph. Here, \( \mathcal{V} \) represents vertices for probing (\( \mathcal{P} \)) and causal (\( \mathcal{C} \)) interventions\footnote{Past probing interventions (\( \mathcal{P}_{<i} \)) likely influence current and future ones (\( \mathcal{P}_i \)), ensuring weak connectivity, and any \( \mathcal{P} \) cannot be the cause of its own \( \mathcal{C} \), thereby guaranteeing acyclicity. This structure reflects the linear progression typical in turn-based dialogues. Potential non-linearities in multimodal contexts~\cite{hunter2018formal} largely do not affect the acyclic structure because multimodal channels tend to overlap rather than invert the linear order of dialogue entirely~\cite{alahverdzhieva2017aligning}.}; edges \( \mathcal{E}_1 \) denotes connectivity between vertices; weights \( \mathcal{E}_2 \)  indicate causal influence from \( \mathcal{C} \) to \( \mathcal{P} \), thereby establishing a total order; and \( \lambda \) is a directed path induction function over \( \mathcal{E}_2 \) and a vertex \( v \in \mathcal{V} \) that emits the root intervention \( \mathcal{C} \) and terminal intervention \( \mathcal{P} \) in \( \mathcal{G} \), implicit in the discourse’s linear order.

\textbf{Definition 2.} Given a deliberation graph \( \mathcal{G} = (\mathcal{V}, \mathcal{E}_1, \mathcal{E}_2, \lambda) \), a \textbf{deliberation chain} (or \textit{intervention cluster}\footnote{We will use {\it deliberation chain} for the ordered sequence of interventions in a dialogue, and {\it intervention cluster} for the clusters output by our system. Both denote a chain of sequential interventions linked by transitive closure, similar to \textit{entity clusters} in coreference literature.}) is a subgraph \( \mathcal{G}' = (\mathcal{V}', \mathcal{E}'_1, \mathcal{E}'_2, \lambda) \) of \( \mathcal{G} \), such that \( \{\mathcal{P}_{\hat{\imath}}, \mathcal{C}_{\hat{\jmath}}\} \subseteq \mathcal{V}' \), where \( \hat{\jmath}  = \min\{ j \mid \lambda(\mathcal{C}_j) \in \mathcal{V}' \} \)  and \( \hat{\imath} = \max\{ i \mid \lambda(\mathcal{P}_i) \in \mathcal{V}' \} \) indicate the initial and final occurrences respectively in the traversal of \( \mathcal{G} \) from $\mathcal{C}_{\hat{\jmath}}$ to  $\mathcal{P}_{\hat{\imath}}$. See Fig.~\ref{fig:full_pipeline}.

We formulate deliberation chain construction as a coreference resolution-style clustering problem \cite{ng2002improving,lee2012joint}, over a dialogue, $D$, with $N$ utterances, that the system must cluster into probing interventions and their linked causes, such that each cluster forms a unique deliberation chain. Given the elements of a cluster, $\lambda$ reconstructs the chain by enforcing transitive closure over the within-cluster links given the temporal order inherent in the discourse, under Definition 1 above. This formulation motivates our joint modeling approach, which is detailed in Sec.~\ref{sec:method}.

\begin{figure}[h!]
    \centering
    \includegraphics[width=.45\textwidth]{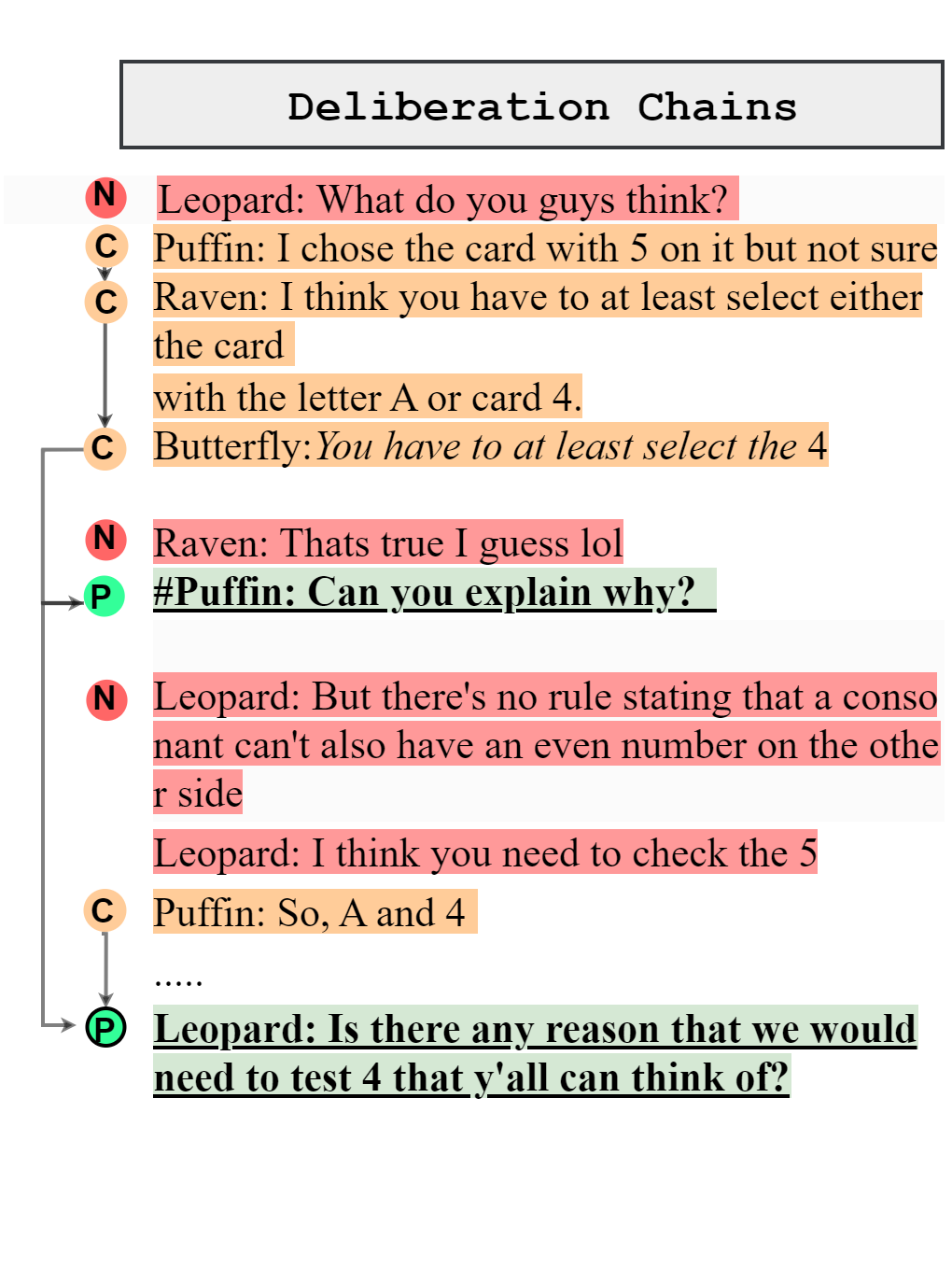}
    \caption{Example of a deliberation chain, showing the flow of interventions and their causal relationships within a collaborative task. This example is adapted from our model’s output on the DeliData corpus.}
    \label{fig:deliberation_example}
\end{figure}

In Fig.~\ref{fig:deliberation_example}, we provide a detailed example of a deliberation chain from our dataset. The causal interventions (\textit{e.g., ``You have to at least select either the letter A or card 4.''}) and probing questions (\textit{e.g., ``Can you explain why?''}) form a structured sequence, where probing interventions are linked to their causal antecedents. This transitive closure forms the deliberation chain, which reflects how participants navigate the problem-solving process.

\section{Dataset Annotation}
\label{sec:data}
\vspace*{-2mm}

We evaluate intervention clustering on two recent, challenging collaborative dialogue datasets: DeliData and the Weights Task Dataset.

\subsection{DeliData}
\vspace*{-1mm}

The DeliData corpus \cite{karadzhov2023delidata} is a publicly-available dataset intended for studying group deliberation in multiparty problem-solving. It comprises 500 group dialogues, totaling 14,003 utterances, centered around the Wason card selection task, a well-established cognitive puzzle~\cite{wason1968reasoning}. Each group contains 5 participants, who are presented with 4 cards that have a number or a letter on them. They must collectively decide which cards to turn over to test the rule, ``All cards with vowels on one side have an even number on the other?'' 
The dataset includes both the dialogues themselves, which denote cards by the symbols on them (letters or numbers), and a measure of decision correctness (task performance) before and after the group discussion, and is annotated with deliberation cues, argumentation structures, and other conversational dynamics. DeliData splits consist of 300, 100, and 100 randomly-chosen groups for training, development, and testing, respectively.


\subsection{Weights Task Dataset}
The Weights Task Dataset (WTD) \cite{khebour2024text} is an anonymized publicly-available dataset intended for studying small group collaboration. It comprises 10 videos, where groups of three participants must use a balance scale to identify the weights of differently-colored weighted blocks and the pattern that describes the weights. The task unfolds in 3 stages, where users solve the problem with the scale, without the scale, and with inferred knowledge of the pattern in weights. The dataset includes multiple annotations, including human gold-standard transcriptions of the participants' dialogues. Utterances reference blocks by color and deduced candidate weights, and can be used to identify probing questions and their potential causal interventions. WTD splits consist of 7, 1, and 2 randomly-chosen groups for  training, development, and testing, respectively.

\subsection{Data Augmentation of WTD}
\vspace*{-1mm}
The WTD is a multimodal dataset, but as the focus of this paper is establishing this novel task, our current study does not incorporate non-verbal cues. Instead, we employ {\it dense paraphrasing} \cite{tu2023dense} as an augmentation technique to explicitly define which blocks are being referred to in the situated dialogue, so that probing and causal interventions can be modeled using just a textual signal. The WTD annotations include dense paraphrased utterances for the first stage but not the second two. We followed the procedure from \citet{khebour2024common} to dense paraphrase the remainder of the dataset (e.g., replacing ``those'' with ``red block and blue block'' in cases where the video makes clear that those blocks are the intended denotata). Utterances were dually annotated (Cohen’s $\kappa$ = 0.69) and adjudicated by an expert.

\begin{figure}[t]
  \includegraphics[width=0.5\textwidth,trim={5px 17px 0 0},clip]{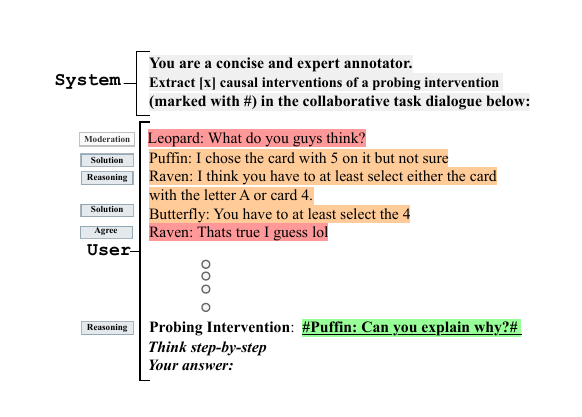}
\vspace*{-2mm} 
\caption{Prompting framework for GPT to select causal interventions given a probing intervention and a dialogue history (example from DeliData). Ground-truth labels for probing and causal interventions are marked in green and brown, respectively.}  \label{fig:probing_prompt}
\vspace*{-5mm}
\end{figure}

\vspace*{-1mm}
\subsection{GPT Annotations of Deliberation Chains}
\vspace*{-1mm}


Like coreference cluster annotation, which often requires exhaustive cross-comparisons across tokens~\cite{bugert2020breaking}, human annotation of deliberation chains is time-consuming and expensive. Therefore, to create ``gold'' chains for fair comparison, we draw on work in LLM-augmented annotations with Chain-of-Thought (COT) reasoning~\cite{radhakrishnan2023question,wei2023chainofthought,nath2024okay} for ``soft'' gold labels. 

We apply a two-pronged strategy.
\begin{enumerate*}[label=(\arabic*)]
    \item We sequentially prompt {\tt GPT-3.5-turbo-0125} using an argument-extraction framework~\cite{ahmed-etal-2024-linear-cross} (see Fig.~\ref{fig:probing_prompt}) to extract causal interventions for all probing interventions in the data\footnote{For the WTD, which does not already contain probing labels, we use the dense paraphrased utterances to extract probing labels before this step. See Appendix~\ref{app:label-gen}.} with prior dialogue history\footnote{A probing intervention can cause another probing statement within a dialogue~\cite{sukmadewi2014improving, behr2012asking}. As such, we do \textit{not} omit probing labels from the previous utterances given as context.} and a system-based task-description to guide its reasoning. We also explicitly ask the LLM to generate free-text rationales (FTRs) corresponding to every causal intervention extracted, to augment its reasoning~\cite{kunz-etal-2022-human, ravi2023happens}.
    \item We do an extensive human evaluation of these LLM-generated annotations to validate quality of extracted clusters. FTRs were used as an additional reference for human evaluators to validate the GPT's annotations and their alignment with human reasoning. This evaluation demonstrated high acceptability of GPT labels and reasoning to humans (see ~\ref{anno:human-eval} for details).
\end{enumerate*}

Since deliberation graphs are weakly-connected, in each iteration we also apply a labeling algorithm (see Algorithm~\ref{alg:gpt_gold_algorithm} in the appendix) to assign the correct preceding cluster for newly appearing interventions in the loop. Table~\ref{tab:cluster_statistics_deli_wtd} provides cluster-level details of the two datasets.

\subsection{Human Evaluation of GPT-Annotated Labels}
\label{anno:human-eval}

We conducted a human evaluation to assess the quality of the GPT-generated annotations on a random representative subset of 25 samples from both DeliData and WTD test sets. These samples were evaluated across several dimensions: relevance, presence in sequence, information sufficiency, acceptability, and rationale overlap. 

\begin{figure}[h!]
    \centering
    \includegraphics[width=.45\textwidth]{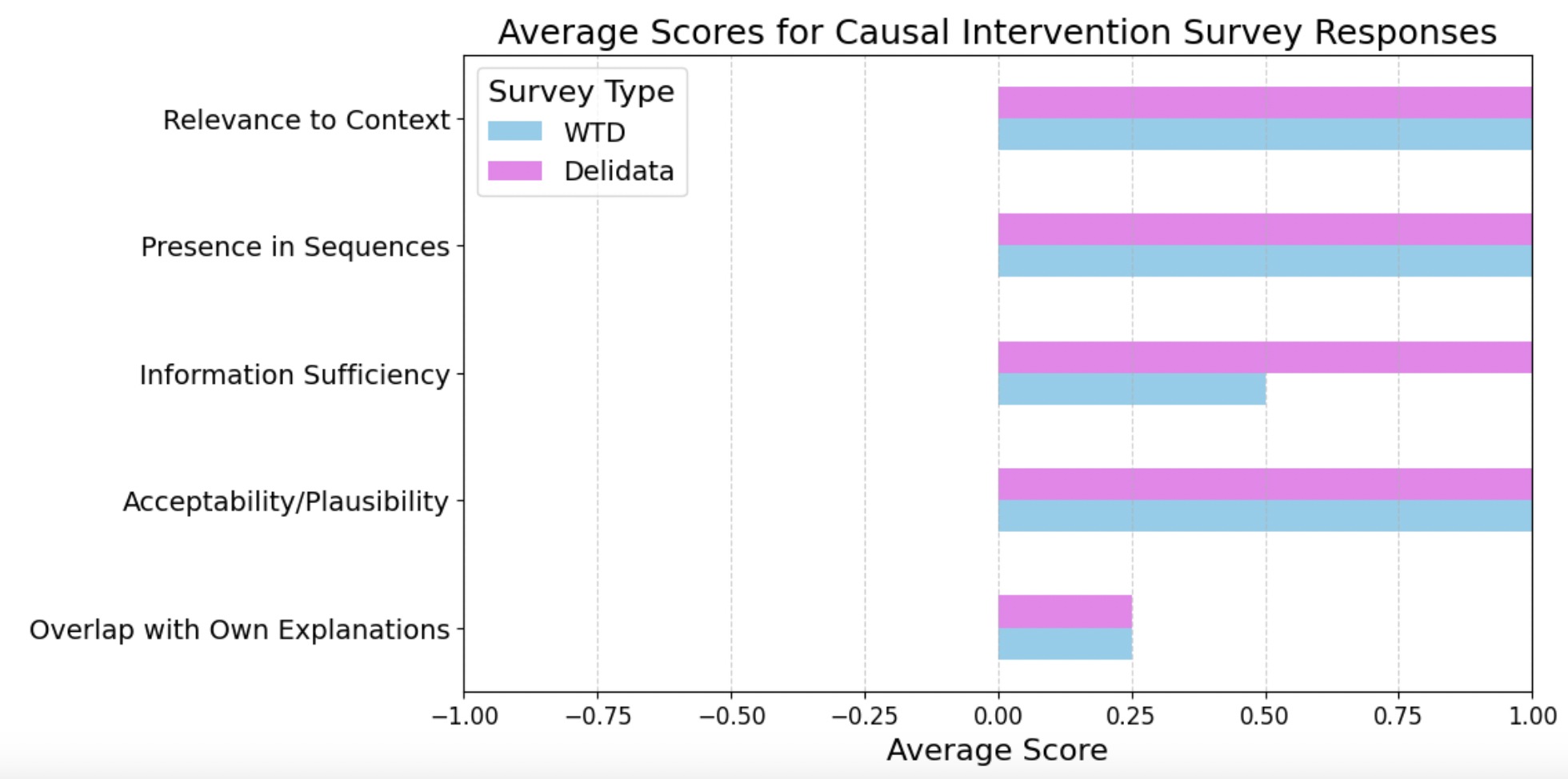}
    \caption{Average Scores for Causal Intervention Survey Responses.}
    \label{fig:data-wtd-deli-eval}
\end{figure}

The annotators consistently agreed that the annotated utterances were indeed causal to the probing utterance, as indicated by high agreement on the first two questions concerning \textit{Relevance to Context} and \textit{Presence in Sequences}. These are the most critical aspects of the evaluation, and the high level of agreement demonstrates that the core annotations were valid. The annotators' answers to questions concerning rationale alignment, however, showed more variability, as expected and seen in Fig.~\ref{fig:data-wtd-deli-eval}. While annotators may agree that an utterance is causal, they may align less with the specifics of the rationale behind why it is causal. This variation is natural and does not impact the overall validity of the annotations.

We calculated Krippendorff’s alpha to measure inter-annotator agreement. Each unique qualitative response was mapped to distinct numerical categories (e.g., Yes, No, Not enough information, Enough information) to capture the differences between responses more effectively. This calculation resulted in Krippendorff's alpha values of 0.88 for DeliData and 0.92 for WTD, indicating strong agreement between annotators on these samples.

Further details on the evaluation process, can be found in Appendix~\ref{app:human-eval}.

\begin{table}[htb]
    \centering
     
    \resizebox{\linewidth}{!}{
    \begin{tabular}{@{}llcccccc@{}}
        \toprule
        & & \multicolumn{3}{c}{DeliData} & \multicolumn{3}{c}{WTD} \\
        \cmidrule(lr){3-5} \cmidrule(lr){6-8}
        & & Train & Dev & Test & Train & Dev & Test \\
        \midrule
        
        \# Probing & & 1005 & 317 & 358 & 115 & 15 & 64 \\
        \# Causal & & 1975 & 637 & 700 & 247 & 37 & 134 \\
        \# Total  & & 2980 & 954 & 1058 & 362 & 52 & 198 \\
       
        Min Chain Length & & 2 & 2 & 2 & 2 & 3 & 2 \\
        Max Chain Length & & 21 & 15 & 14 & 14 & 12 & 13 \\
        Mean Chain Length & & 5.4 & 5.5 & 5.1 & 5.1 & 5.2 & 4.7 \\
       \# Clusters & & 552 & 174 & 206 & 71 & 10 & 42\\
       Avg. Dialogue Length   & & 33 & 35.3 & 34.9 & 222 & 204 & 326\\
       Tokens to Probing & & 227 & 241 & 214 & 293 & 367 & 276 \\

        \bottomrule
    \end{tabular}
    }
    \vspace*{-2mm}
    \caption{Descriptive statistics of true (gold) intervention clusters/deliberation chains in DeliData and Weights Task datasets across different splits. Note that there are no singletons in either dataset. ``Length'' refers to utterances. Last row denotes mean number of tokens from start of a dialogue until a probing intervention.}
    \label{tab:cluster_statistics_deli_wtd}
    \vspace*{-2mm}
\end{table}



\section{Joint Learning of Deliberation Chains}
\label{sec:method}
\vspace*{-2mm}

To automatically cluster interventions that form a deliberation chain \( \mathcal{G}'\), a model must learn to assign, for each possible $\mathcal{P}_i$, the most suitable antecedent utterance $\mathcal{C}_j$, that forms a correct link in the chain. Prior works in coreference resolution~\cite{lee-etal-2017-end, zhang2018neural} typically addressed such assignments using joint-learning frameworks that exhaustively score antecedent ``spans'' and thereby produce coreference chains. Our approach implicitly produces the correct chain since interventions in a dialogue follow a linear order assuming transitivity across links. 

Standard joint-learning frameworks for coreference resolution
typically operate at the \textit{span}-level. For our task, where the entire deliberative utterance forms a distinct discourse unit~\cite{hunter2018formal}, this is an incompatible approach. As such, we propose a joint-learning framework that models the task as a conditional probability distribution \(Pr(P, C, L \mid D)\), partitioned into multinomial probabilities, assuming that utterance spans are conditionally independent given the dialogue $D$. Mathematically,
 
\vspace{-4mm}
\begin{equation}
\small
\resizebox{\columnwidth}{!}{
$Pr(P, C, L \mid D) = \prod_{i=1}^{N} \prod_{j=1}^{N} Pr(p_i \mid D) Pr(c_j \mid D)  Pr(l_{ij} \mid D)$,}
\label{eq:main_probability}
\end{equation}
where $P$ refers to the probability of an utterance being a Probing intervention, $C$
 refers to the probability of an utterance being a Causal intervention, and $L$ 
 refers to the probability of a Link between the two utterances. $p_i$, $c_j$ and $l_{ij}$ are treated as random variables denoting the probabilities of an utterance being probing, being causal, and of the link between the two interventions, respectively; $N$ denotes the number of individual utterances within a dialogue $D$. 






\subsection{Model}
\label{ssec:model}
\vspace*{-1mm}

\paragraph{Intervention Pair Representation}

As the right-hand side of Eq.~\ref{eq:main_probability} represents causal dynamics as probabilities of links between pairs of utterances in the discourse, we draw on a cross-encoding strategy from coreference research~\cite{humeau2020poly, cattan2021cross, ahmed20232} to score pairs of utterances. Since some dialogues, especially in the Weights Task Dataset, can reach up to $\sim$200 utterances, we use the Longformer model~\cite{beltagy2020longformer} as the base encoder. To construct an expressive representation for a pair of interventions ($\mathcal{P}_i$ , $\mathcal{C}_j$), we first demarcate their start and end with special tokens ({\tt <m>} and {\tt </m>}). For context around a probing intervention, we also concatenate the $k$ previous utterances\footnote{Setting $k=10$ and max sequence length (probing intervention with preceding utterances) to 512 was empirically found to cross-encode both utterances in a pair, on average, without losing expressive tokens or incurring inordinate compute cost. See Table~\ref{tab:cluster_statistics_deli_wtd} for more details.}
 along with participant name or number as given in the dataset. We extract the {\tt [CLS]} token representation of this concatenated input, the cross-attentional context of $\mathcal{P}_i$ and $\mathcal{C}_j$, as well as their Hadamard product, $\mathcal{P}_i \odot \mathcal{C}_j$. 
This results in a combined vector representation for pair ($\mathcal{P}_i$, $\mathcal{C}_j$):
\begin{equation}
    V(\mathcal{P}_i, \mathcal{C}_j) = [V_{CLS}, V_{\mathcal{P}_i}, V_{\mathcal{C}_j}, V_{\mathcal{P}_i} \odot V_{\mathcal{C}_j}]  
    \label{eq:combined_embedding}
\end{equation}

Next, to maximize the log-likelihood in our joint-learning framework (Eq. \ref{eq:main_probability}), we generate three sets of scores from specific segments of Eq. \ref{eq:combined_embedding} using three feed-forward neural networks (FFNN): 
\begin{enumerate*}[label=(\arabic*)]
    \item a linking score $l_{ij} = {\text{FFNN}_l(V(\mathcal{P}_i, \mathcal{C}_j))}$, the probability of a pair of utterances forming a true link; and
    \item two intervention scores, $s_{i} = {\text{FFNN}_p(V_{\mathcal{P}_i})}$ and $s_{j} = {\text{FFNN}_c(V_{\mathcal{C}_j})}$ of the candidate and the antecedent, respectively, being valid interventions. 
\end{enumerate*}


\begin{figure*}[t]
  \centering
  \includegraphics[width=\textwidth]{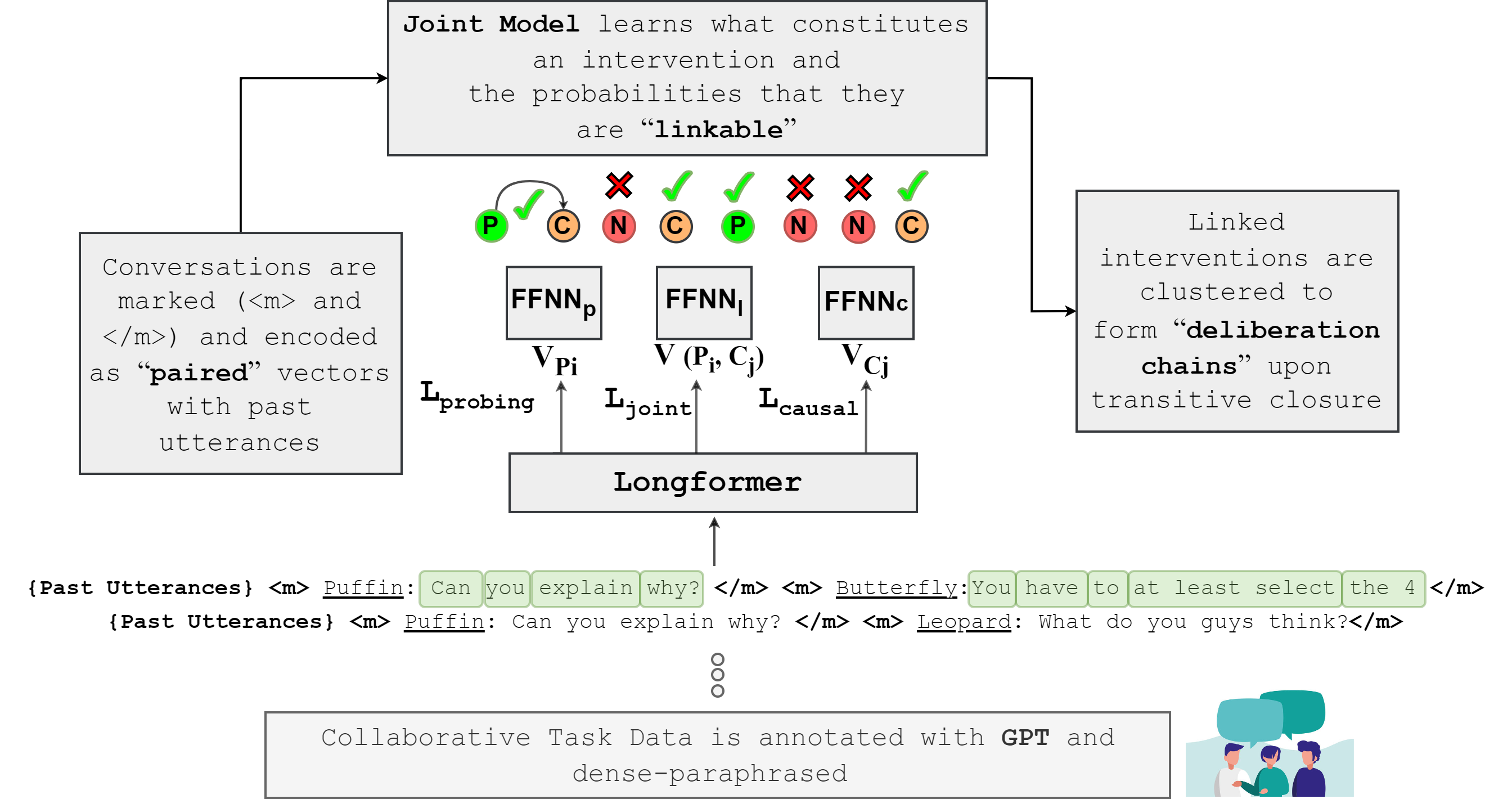}
  \caption{Our joint-learning framework for \textit{deliberation chains}, learning to assign correct antecedent utterances for every valid intervention using a ``probing'' score, a ``causal'' score, and a ``linking'' score. Pairs of utterances are encoded with global attention (in green between {\tt <m>} and {\tt </m>}), further contextualized by past utterances.}
  \label{fig:full_pipeline}
\end{figure*}

Thus, the model picks up on two types of learning signals: correctly assigning a true antecedent to a candidate intervention while also learning what constitutes a valid intervention. We directly optimize the model with $\mathcal{L}_{\text{joint}}$:

\vspace*{-4mm}
\begin{equation}
    \mathcal{L}_{\text{joint}} =  \mathcal{\alpha}_{\text{p}} \mathcal{L}_{\text{probing}}  +  \mathcal{\alpha}_{\text{c}} \mathcal{L}_{\text{causal}}  + \mathcal{\alpha}_{\text{l}} \mathcal{L}_{\text{link}} 
\end{equation} that consists of a weighted-combination of three separate loss terms. $\mathcal{L}_{\text{probing}}$ and $\mathcal{L}_{\text{causal}}$ are each defined as:

\vspace*{-4mm}
\begin{equation}
    \mathcal{L}_{\text{[probing,causal]}}(*) = -\sum_{*=1}^{N} y_* \log(\sigma({s}_*))
\end{equation}
where $*$ corresponds to $i$ and $j$ in $\mathcal{L}_{\text{probing}}$ and $\mathcal{L}_{\text{causal}}$, respectively, $\sigma$ is the sigmoid function, and $y$ is the predicted output. The final term is $\mathcal{L}_{\text{link}}$:




\vspace*{-4mm}
\begin{equation}
\small
 \mathcal{L}_{\text{link}}(i,j) = - \sum_{i=1}^{N}\sum_{j=1}^{N} y_{ij}\log ({l}_{ij}) +  (1- y_{ij})\log (1-{l}_{ij})
\end{equation} 

See Fig.~\ref{fig:full_pipeline} for further details.
\(\mathcal{\alpha}_{\text{p}}\), \(\mathcal{\alpha}_{\text{c}}\) and \(\mathcal{\alpha}_{\text{l}}\) are learned regularization parameters tuned on the development sets of our data. Following~\citet{nath2024okay}, we fixed \(\mathcal{\alpha}_{\text{l}}\) = 1 and \(\mathcal{\alpha}_{\text{c}}\), \(\mathcal{\alpha}_{\text{p}}\) = 0.01 after initial experiments.




\vspace*{-1mm}
\paragraph{Training Pair Generation}

For training a pairwise scorer model, an efficient pair generation process is crucial. A naive way to implement Eq.~\ref{eq:main_probability} compares each utterance \( u_i \) to the set of all its preceding antecedents \( U(i) = \{\epsilon, \ u_1, \ldots, u_{i-1}\} \) to generate pairwise scores.\footnote{Our training method is generalizable to all utterances, since the ground truth label on any candidate can be causal, probing, or \textit{neither} (a non-intervention dummy variable, $\epsilon$). Generated pairs may have true labels that are any combination of probing and causal, since two causal interventions may be linked to the same probing intervention, or two probing interventions may share a cause, which results in these pairs themselves being linked under transitive closure. This follows standard practice in pairwise approaches to coreference across long documents. 
} This results in \( \sim O(N^2) \) complexity for a dialogue of $N$ utterances. Discourse-coherence theory~\cite{grosz1986attention, held-etal-2021-focus} suggests that the most pertinent information to a specific utterance remain \textit{within} an ``attentional state'', i.e., the point of focus of participants within a dialogue. As such, given a dialogue of $N$ utterances, for each target $u_i$, we define a window $W$ of previous utterances considered for training. Because of the long tail of true negative samples (non-links), this value is tuned over the dev split of each dataset to make the ratio of positive to negative samples more balanced (cf.~\citet{ahmed20232} for optimal training.). Given a true intervention cluster after annotation and labeling, all pairs within it are considered positive pairs. Negatives comprise all other pairs under consideration (which may be limited by window $W$). 

During training, the model is forced to learn discourse-relevant signals from the positive pairs drawn from true intervention clusters.
Applying Longformer's \textit{global attention} to \textit{all} tokens in the pair (Fig.~\ref{fig:full_pipeline}) allows us to encode relevant global features within $W$.
Utterances in the preceding neighborhood $W$ typically display lexical overlap for items with similar semantic roles, or task-specific phrases.\footnote{For instance, in the Weights Task, neighboring utterances contain overlapping arguments like ``red block'' when the group is solving a particular subtask relevant to that block.} 
When such pairs are sourced from separate intervention clusters that occur within $W$, they naturally form difficult samples for encoder-only LLMs like Longformer due to misleading lexical overlap~\cite{ravi2023happens,ahmed20232}.

\vspace*{-1mm}
\paragraph{Inference}
We evaluate two inference strategies. For our naive approach, we relax $W$ and generate all candidate antecedent utterances within $D$, score them using using the intervention scores (mean of $s_i$ and $s_j$), and only keep the remaining pairs based on a threshold $\tau$ (details in Appendix~\ref{app:pruning}). This reduces cross-comparisons in building the intervention clusters as we only use the pairwise scorer to score the remaining utterances. We also consider all scores generated without relaxing $W$. While the naive approach tests the system's recall under a long-tail of true-negatives, this method enforces a more balanced distribution, resulting in a ``soft'' upper-bound on model precision. Pairwise scoring generates an adjacency matrix of links between utterances. Inducing transitivity between links using a connected-components based clustering method with a threshold of 0.5 generates the final intervention clusters. Under temporal ordering, these expand to deliberation chains within a dialogue.

\vspace*{-2mm}
\section{Experiments}
\label{sec:exp}
\vspace*{-2mm}

We evaluated our joint modeling method against 3 similarity baselines and two cross-encoder methods adapted from coreference research.

\vspace*{-1mm}
\subsection{Similarity Baselines}
\vspace*{-1mm}
    
For simple similarity baselines, we assessed:

\begin{itemize}
    \vspace*{-2mm}
    \item Simple token overlap between utterances. This may indicate correspondence between a probing intervention and its cause(s), as the utterances may share terms. To assess lexical similarity between utterance pairs, we calculated the Levenshtein distance ratio (0--100) between the two strings.
    \vspace*{-2mm}
    \item The overlap of salient \textit{entities} within the utterances. We computed an Intersection over Union (IoU) of entity counts score based on categorical features derived from task-relevant categories referenced in each utterance (i.e., vowels, consonants, even and odd numbers in DeliData, and colors and weights in WTD).
    \vspace*{-2mm}
    \item Cosine similarities between embeddings of the two utterances, extracted from BERT-{\tt base-uncased}, following the intuition that probing utterances should share some \textit{semantic}, not just token or entity similarity~\cite{jawahar2019does} with their causal counterparts.
    \vspace*{-2mm}
\end{itemize}

For each, we set a threshold value for each dataset, equal to the average of the relevant metric over the dev set. If the relevant metric for a test pair exceeded this calculated threshold for the dataset, we linked that pair. More details on these baselines and the threshold values are given in Appendix~\ref{app:baselines}.

\vspace*{-2mm}
\subsection{Cross-Encoder Baselines}
\vspace*{-1mm}
For trainable baselines, we specifically chose recent coreference resolution frameworks that operate on an ``utterance'' level (instead of a span-level) for a valid comparison (see Sec.~\ref{sec:method}). For fairness, we used the base encoders from these frameworks as well as with their cross-encoding strategies, but not their fine-tuned weights, since fine-tuning on a separate task can likely tilt the model out of distribution~\cite{kumar2022finetuning}.   

We used \citet{caciularu2021cdlm}'s Cross-Document Language Model (CDLM). with a context length of 1,024 preceding tokens along with their cross-encoding setup.\footnote{CDLM (\url{https://huggingface.co/biu-nlp/cdlm}) is trained on documents with overlapping information and is suitable for handling long inputs, which are both traits of our dialogues (Sec.~\ref{ssec:model}). For compute reasons, we truncate pairs at a maximum sequence length of 1,024 tokens after tokenization since the token-length of utterance pairs in training is $\sim$220 tokens for both datasets, on average.} We also employed \citet{ahmed20232}’s ``bidirectional'' BCE loss-based learning method. This generates a mean of the BCE losses over the forward pass of utterances paired in both directions: ($u_i$, $u_j$ and $u_j$, $u_i$). Like our joint modeling approach, the context window here is 512 tokens.

\vspace*{-2mm}
\subsection{Joint Modeling Hyperparameters}
\vspace*{-1mm}

For joint modeling, we use the Adam~\cite{kingma2014adam} optimizer with batch size 24, with learning rates of $1e-6$ for the encoder fine-tuning, $1e-4$ for the pairwise scorers, and $1e-5$ for the intervention scorers. Each training epoch on an NVIDIA A100 took $\sim$20 and $\sim$40 minutes for DeliData and WTD, respectively. Each model was evaluated after a single training run for 16 epochs after robust hyperparameter tuning on the validation sets. 

\section{Results}
\label{sec:results}
\vspace*{-2mm}

\begin{table*}[!ht]
    \centering
     
 \footnotesize
    \begin{tabular}{@{}lllrrrrrrrrrrrrrrr@{}}
    \toprule
    &&& \multicolumn{5}{@{}c@{}}{DeliData} && \multicolumn{5}{@{}c@{}}{WTD} \\
    \cmidrule{4-8} \cmidrule{10-14}
    &&& \multicolumn{3}{@{}c@{}}{$B^3$} && CoNLL && \multicolumn{3}{@{}c@{}}{$B^3$} && CoNLL\\
    \cmidrule{4-6} \cmidrule{8-8} \cmidrule{10-12} \cmidrule{14-14}
    &&& R &P & $F_1$ && \multicolumn{1}{r}{$F_1$} && R &P & $F_1$ && \multicolumn{1}{r}{$F_1$} \\ 
   \midrule
       & Lexical Overlap && 26.6 & 81.3 & 40.0 && 28.6 && 41.6 & 50.0 & 45.4 && 36.6 \\
       & Entity Overlap  && 34.9 & 71.7 & 46.9 && 40.6 && 27.2 & 70.0 & 39.2 && 26.7 \\
       & BERT-Cosine     && 98.6 & 49.9 & 66.3 && 69.2 && \textbf{100.0} & 7.1 & 13.2 && 35.3 \\
       & LongContext     && 84.7 & 60.7 & 70.7 && 68.2 && 72.1 & 23.8 & 35.8 && 45.5 \\
       & Bidirectional   && 90.8 & 59.2 & 71.7 && 70.9 && 64.5 & 31.5 & 42.4 && 44.3 \\
       & LLaMA 2-7B-chat && \textbf{99.9} & 49.7 & 66.4 && 69.7 && \textbf{100.0} & 7.1 & 13.2 && 35.3 \\
       \cdashline{1-14}
       & --- Ours (Joint - $W$) && 92.3 & 60.5 & 73.1 && 73.6 && 54.4 & \textbf{75.0} & 63.0 && 50.3 \\
       & --- Ours (Joint + $W$) && 87.8 & \textbf{72.6} & \textbf{79.5} && \textbf{76.4} && 67.9 & 61.7 & \textbf{64.7} && \textbf{58.1} \\

    \bottomrule
    \end{tabular}
    \vspace*{-2mm}
    \caption{$B^3$ and CoNLL $F_1$ metrics on DeliData and WTD test set results. ``LongContext'' denotes \citet{caciularu2021cdlm}'s coreference methodology applied to deliberation chain clustering. ``Bidirectional'' denotes \citet{ahmed20232}'s methodology.}
    \label{tab:abbr_results}
    \vspace*{-2mm}
\end{table*}

We evaluate against coreference methodology using cluster metrics computed using the CoVal coreference scorer~\cite{moosavi2019minimum}, specifically $B^3$ and CoNLL $F_1$ metrics, as presented for both datasets in Table~\ref{tab:abbr_results}.\footnote{Since we are using the gold intervention labels for our experiments, using $B^3$ is more reliable compared to other metrics~\cite{moosavi2016coreference, held-etal-2021-focus}.} We also present zero-shot results from LLaMA 2-7B-chat. The prompting framework for this is given in Appendix~\ref{app:zero-shot}. Appendix~\ref{app:addl-results} presents results according to other common coreference metrics. Results empirically demonstrate the strength of our theoretically-grounded method on this challenging task and data. 

The multimodal nature of the WTD likely makes it more challenging than DeliData due to cues that may be missed in even the dense paraphrased language. 
The use of the windowed approach results in a small performance improvement due to the exclusion of false positive links outside $W$. The BERT-Cosine and LLaMA 2 zero-shot baselines perform extremely similarly (returning identical metric values on WTD) and achieve perfect or near-perfect recall. This is likely due to these methods returning a very high proportion of false positive links and transitive closure subsequently clustering (almost) all interventions in a dialogue.

\section{Discussion}
\vspace*{-2mm}
\label{sec:disc}

\begin{table*}[h]
\resizebox{\textwidth}{!}{
    \footnotesize  
    
    \begin{tabular}{lp{10.5cm}p{8cm}}
    \toprule
    & \textbf{Dialogue} & \textbf{Free-Text Rationale(s)} \\
    \midrule
    (a) & {\bf [C1] Emu: I picked the card with the vowel A on it, because the rule said all cards with vowels on one side will have an even number on the other} & [C1] \textit{"Emu's statement directly relates to the reasoning behind choosing the card with the vowel A, which is crucial in the} \\
    & [C2] Koala: I think it is A and 2 & \textit{decision-making process."}\\
    & [C3] Hamster: I agree & \\
    & ... & \\
    & {\bf [P4] Bee: So are we ready to final submit} & \\
    
    \midrule
    
    (b) & [C1] Narwhal: What card did you think needed to be turned? & [C3] \textit{"This statement hints at the strategy of testing a card that} \\
    & ... & \textit{would break the rule to confirm its validity, indicating a shift in}\\
    & [C2] Guinea pig: I picked 6 and U & \textit{the participant's thought process during the discussion."} \\
    & ... \\
    & {\bf [C3] Kiwi: We need to pick one that wouldn't fit the rule to test it. Maybe?} \\
    & ... \\
    & {\bf [P4] Kiwi: 7 and U?} \\
    
    \midrule
    
    (c) & [C1] Participant 2: Oh maybe I'll try holding it here & [C2] \textit{"This utterance indicates the participant's initial attempts}\\
    & ... & \textit{to compare the weights of various blocks using their fingers,} \\
    & \textbf{[C2] Participant 2: Mystery block, blue block, red block, green block, purple block, yellow block kinda feels the same} & \textit{setting the groundwork for exploring different measurement techniques."} \\
    & ... & \\
    & \textbf{[C3] Participant 1: So how about purple block, green block two, I had eh purple block, yellow block two} & [C3] \textit{"This utterance directly led to the probing question as it involved a new approach of grouping blocks on fingers to} \\
    & ... & \textit{measure their weights."} \\
    & \textbf{[P4] Participant 2: Is there a better way to measure mystery block?} & \\

    

    \midrule
    
    \bottomrule
    \end{tabular}}
    \vspace*{-2mm}
    \caption{Test samples from DeliData (a \& b) and WTD (c). Bolded utterances indicate $(\mathcal{P},\mathcal{C})$ pairs that our method (Joint - $W$) linked correctly and all other methods failed to. FTRs are given for the annotation of the indicated utterance as causal. These are not included in the input for inference, but are provided as indicators of the kinds of information our framework is likely to learn from the labels that were created using this COT-guided process.}
     \label{tab:error-analysis}
    \vspace*{-2mm}
\end{table*}

\paragraph{Quantitative Analysis}

\begin{figure}[t]
  \includegraphics[width=0.5\textwidth]{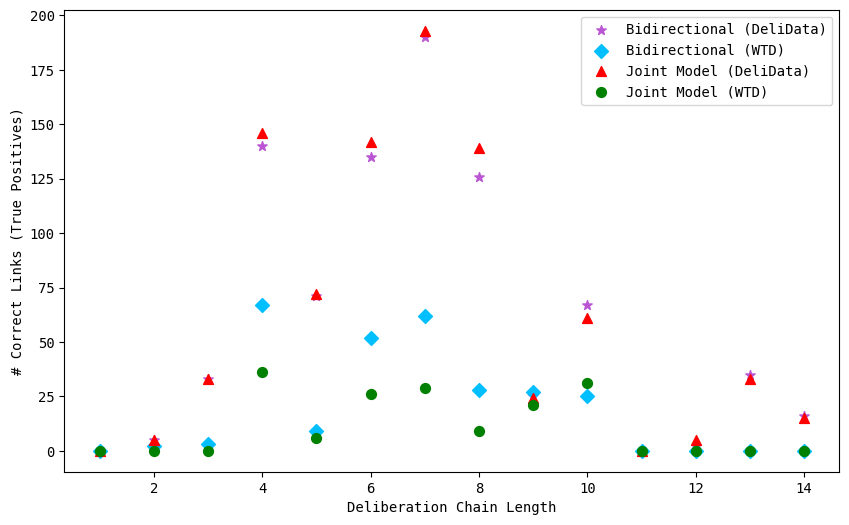}
    \vspace*{-2mm}
\caption{Cluster-level distribution of correctly assigned intervention links for the best-performing cross-encoder baseline compared to Joint - $W$ on both datasets.}  \label{fig:cluster_analysis}
    \vspace*{-2mm}
\end{figure} 

Fig.~\ref{fig:cluster_analysis} shows the count of correct links between interventions assigned by the bidirectional baseline and our (non-windowed) joint model for each cluster size.\footnote{Only the non-windowed model results in a full comparison to all other baselines because Joint + $W$ does not consider all gold pairs.} DeliData has much longer chains on average with a wider distribution at every chain-size, and the joint model consistently links more pairs correctly in frequent medium cluster sizes, while at larger cluster sizes joint modeling and the bidirectional baseline are competitive. The joint model may be learning a more global representation of deliberation chains, as from a discourse-coherence perspective mid-sized chains better reflect the true distribution in collaborative dialogues.

In WTD, the distribution of cluster sizes is narrower, while dialogues are much longer. Our joint model links interventions more conservatively, but also more correctly, than the bidirectional model. This is most evident in the joint model's $\sim$45-point increase in $B^3$ precision compared to the bidirectional baseline. The latter's aggressive linking, while boosting recall for mid-sized clusters, does so at a higher cost to precision. This suggests that when the cluster distribution is skewed (smaller chains, longer dialogues), the joint model is better at avoiding impure clusters.

\vspace*{-1mm}
\paragraph{Qualitative Analysis}
Table~\ref{tab:error-analysis} presents two test pairs from each dataset that our non-windowed model classified successfully that all other methods did not. Utterances are numbered and labeled as ``C'' (causal) or ``P'' (probing). For space reasons, examples given at the link level, not the full cluster level, but we include some other utterances that are clustered into the same chain, as well as the FTR generated by GPT during the COT-guided intervention labeling process, for context and to illustrate the kind of information our method is able to leverage for its decisions that others cannot:

\begin{enumerate}[label=(\alph*)]
    \vspace*{-1mm}
    \item Our model links P4 to C1, which references the letter A, vowels, and even numbers, which are also referenced in C2, which states what the participants agree on. P4 elicits confirmation of all that aggregate information.
    \vspace*{-1mm}
    \item We see in the FTR that C3 and P4 indicate a shift in Kiwi's thought process, and our system picks up the link between the causal and probing utterances made by the same participant, which others miss.
    \vspace*{-1mm}
    \item In this example from the WTD, our model makes two links (between P4 and C2 and C3). Both C2 and C3 pertain to measurement techniques but this is not immediately apparent from the text. The FTR makes apparent that the GPT labels are based on the probing utterance's mention of measuring blocks. 
\end{enumerate}
\vspace*{-2mm}

We note that our model tends to successfully make links much further back in the dialogue history than even the longer-context CDLM model. In the examples presented, we show only the causal and probing interventions that form the response cluster, omitting utterances that are neither (indicated by ellipses). It is notable that these utterances alone still form complete exchanges.

\section{Conclusion and Future Work}
\label{sec:conc}
\vspace*{-2mm}

In this paper, we established a novel task of {\it deliberation chain construction} in collaborative dialogues. We developed a formal graphical model of deliberation chains grounded in discourse coherence theory, and applied coreference resolution techniques to two challenging datasets. Our joint modeling approach emerged as the best model on both datasets, setting a performance standard in this novel task.

Our joint model predicts the probability of an utterance being probing or causal and uses only prior context---a next logical step is to adapt our method to a live interaction, doing intervention \textit{detection}, and predicting when a probing utterance will (or should) occur. This would represent a significant step forward for AI systems that can mediate real-time collaboration.


The WTD's multimodal aspect represents a rich opportunity to investigate multimodality's role in deliberation. For instance, a gesture or action might itself be a probing or causal intervention, and \citet{asher2020modelling} provide a compatible multimodal framework in which to pursue this.

Finally, deliberation chains construction is adaptable to interactions with particular characteristics, like argumentation \cite{afantenos2014counter}, and computational understanding these distinctions will open new horizons in human-AI interaction.

\section*{Limitations}
\vspace*{-2mm}

We used {\tt GPT-3.5-turbo-0125} to annotate utterances as causal interventions given a corresponding probing intervention. While machine-assisted annotation is an accepted method in the field~\cite{vossen2018don,ahmed-etal-2024-linear-cross} and we validated the annotations with human judgments (see Appendix~\ref{app:human-eval}), there is always a risk that annotations provided by AI are noisy or unreliable. Therefore, one limitation that could be addressed in future work is the lack of true gold-standard human annotations for probing and causal interventions in the two datasets.
Additionally, we believe future work should directly compare human vs. automated annotations of things like speech transcriptions in datasets like the WTD in order to further validate the use of automated annotations on these datasets. Finally, we noted that utterances which prompted actions---such as a participant instructing another to move a block---were marked as probing when we used GPT to annotate probing utterances in the WTD. This created a type of probing label that was not present in the DeliData, which entails no physical actions. This may be an indication of the type of noise introduced by automated labeling, but we also speculate that this type of probing is a potential avenue for investigating dialogue driven tasks that require action---future work will need to investigate the validity of such annotations.

Since we adapt coreference techniques to deliberation chain construction, we also use coreference metrics.  \citet{moosavi2016coreference} note the different pitfalls of all common coreference metrics---however, future work should examine the specific limitations of these metrics in the context of the task; some of these metrics may not be a good fit for this and new metrics may need to be developed. See Appendix~\ref{app:addl-results} for a more detailed look into our current thoughts on this limitation.

\section*{Ethical Statement}
\vspace*{-2mm}

Perhaps the largest risk inherent in systems that model deliberation and probing is how they are deployed. Consider, for example, a classroom context: modeling deliberation in a normative fashion may risk disadvantaging persons whose modes of collaboration are non-normative. In the worst case, a system could identify these persons as lacking engagement or having poor collaboration skills, resulting in undue punitive measures.

These models are designed to monitor and aid interaction; however, we do not believe such systems should exist in isolation---explicitly, in a classroom context, we believe such systems should augment teachers, not replace them.

Especially for multimodal use cases, like the Weights Task Dataset (WTD)~\cite{khebour2024text}, there is a risk of such technologies being used for tracking and surveillance, as modeling how individuals collaborate also involves modeling their linguistic and reasoning patterns, which may be sensitive. In this paper, we use publicly-available anonymized datasets that were collected under protocols reviewed by institutional review boards for ethical research, and were conducted with subjects who consented to the release of the data. However, collaboration modeling technology should be treated cautiously when it comes to ingesting multiple modal channels about specific people.

Finally, extending the model of deliberation to an agent that actually intervenes in dialogues could be exploited by bad actors who create bad agents, that bring dialogue to a halt through excessive introduction of ``friction,'' thus impeding the reasoning and productive benefits that collaboration brings. 

These risks are inherent in the deployment of systems that perform the task we have developed herein as a precondition for other actions in the world, not the formal or computational model of deliberation itself.

\section*{Acknowledgements}
This material is based in part upon work supported by Other Transaction award HR00112490377 from the U.S. Defense Advanced Research Projects Agency (DARPA) Friction for Accountability in Conversational Transactions (FACT) program, and by the National Science Foundation (NSF) under a subcontract to Colorado State University on award DRL 2019805 (Institute for Student-AI Teaming). Approved for public release, distribution unlimited. Views expressed herein do not reflect the policy or position of, the Department of Defense, the National Science Foundation, or the U.S. Government. All errors are the responsibility of the authors. Our thanks also go out to the anonymous reviewers whose feedback helped improve the final copy of this paper, and to Maniteja Vallala, Shamitha Gowra, Tarun Varma Buddaraju, and Sai Kiran Ganesh Kumar for extensive data annotation. We also thank Bruce Draper for the lively discussions that contributed to this work.
\bibliography{custom, coreference_reasoning, reasoning}

\appendix

\section{Performance of Linking vs. Turns Between Interventions}
\label{app:performance_turn_length_between_interventions}
\vspace*{-2mm}

\begin{figure}[h!]
    \centering
    \includegraphics[width=\linewidth]{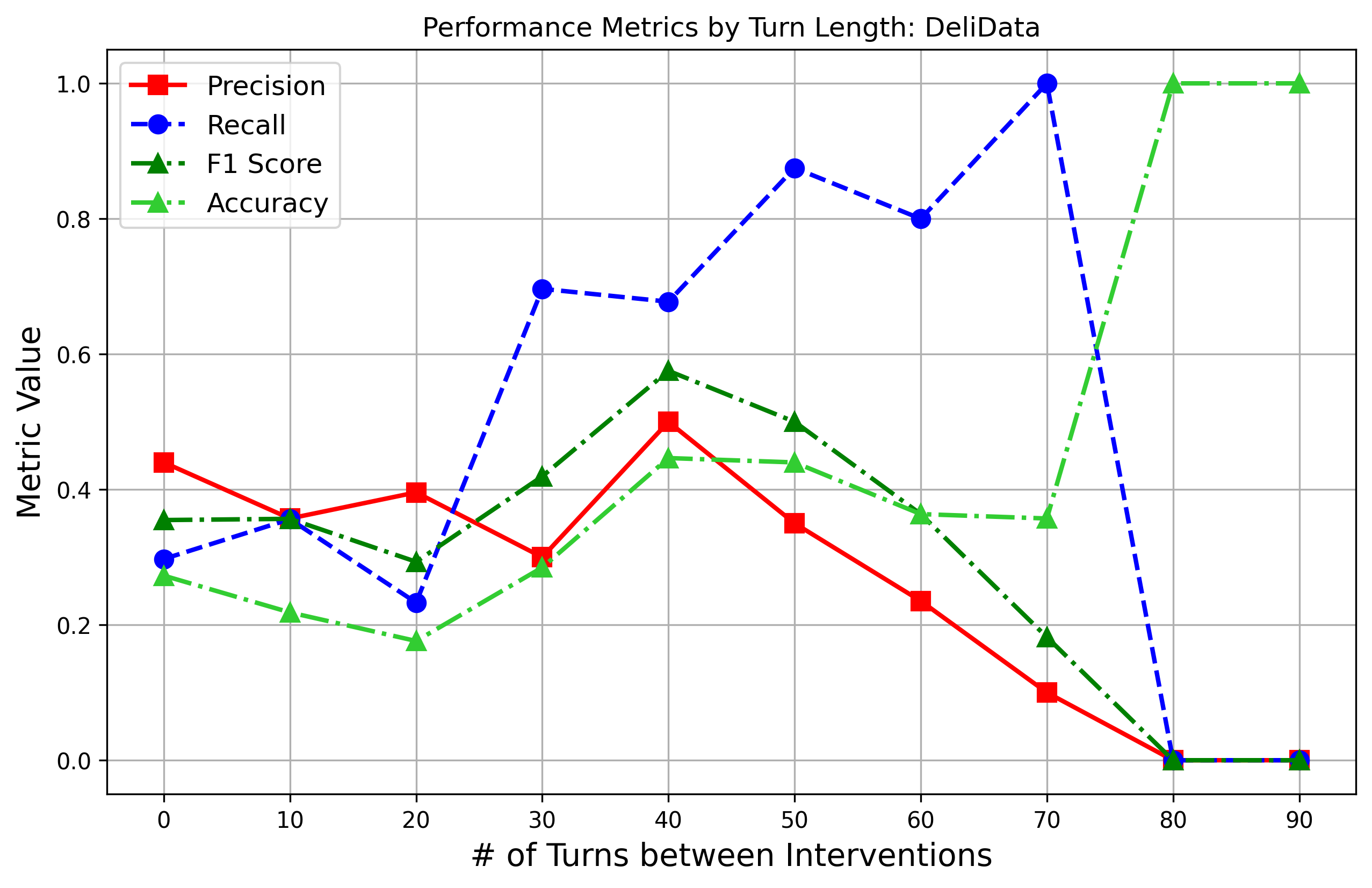}
\vspace*{-2mm}
    \caption{Ablations of our Joint + $W$ model on DeliData by number of turns between interventions, when run on all possible intervention pairs.}
    \label{fig:turn_length_deli}
\vspace*{-2mm}
\end{figure}

\begin{figure}[h!]
    \centering
    \includegraphics[width=\linewidth]{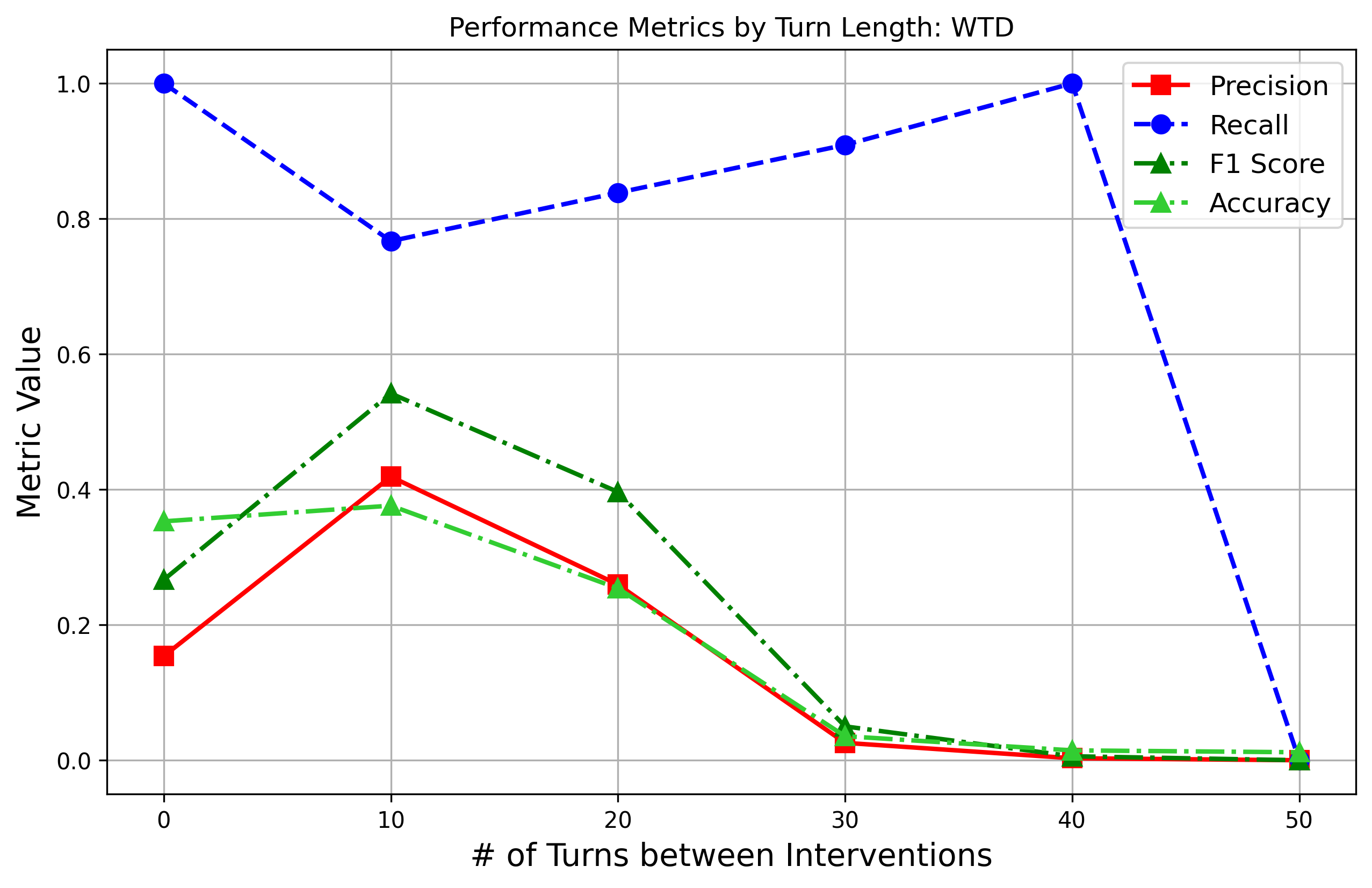}
\vspace*{-2mm}
    \caption{Ablations of our Joint + $W$ model on the Weights Task Dataset by number of turns between interventions, when run on all possible intervention pairs.}
    \label{fig:wtd-turn_length_wtd}
\vspace*{-2mm}
\end{figure}

For both training and inference for the Joint + $W$ model, we fix the value of $W$, the window of previous utterances considered during pair generation, after tuning on the development set (18 for DeliData and 9 for WTD). This establishes a soft-upper limit on the model’s performance since the model only learns from a balanced distribution of pairs likely representing an ``attentional state''~\cite{grosz1977representation, held-etal-2021-focus}. To specifically evaluate this model on pairs that appear are widely separated in the discourse, we conduct ablations considering a range of turn-lengths. We bin all pairs based on their relative distance in the dialogue and report linking performance of the Joint + $W$ model at a pair-level within the bins to provide further insights into its behavior. 

As seen in Fig.~\ref{fig:turn_length_deli} for DeliData, we find that overall precision dips slightly as turn-length increases and reaches a maximum within pairs binned between 30 and 40, after which it declines until all pairs are exhausted. Similar peaks for precision are also seen on WTD (Fig.~\ref{fig:wtd-turn_length_wtd}), albeit at a shorter turn-length of 10 utterances. This suggests that our joint learning framework is likely helping the model pick up signals of the ``validity'' of interventions \textit{per se}, without having to solely rely on previous context as in its encoding strategy (since we restrict $k$ to be 10). On the other hand, the much smaller size of the WTD corpus with a narrower distribution of clusters and much longer dialogues makes it more challenging for our model to assign correct links to widely separated interventions in the discourse. 

\section{Pruning Pairs for Naive Approach}
\label{app:pruning}

Since our naive approach while relaxing $W$ (Sec.~\ref{ssec:model}) still operates within an utterance-level pairing of interventions, unlike ``span''-level pruning strategies, our threshold $\tau$ directly prunes at the utterance-pair level instead of using a token-based filtering to improve recall~\cite{cattan2021cross}. Specifically, once all antecedents have been scored using the two intervention FFNNs (mean of $s_i$ and $s_j$), we retain the top $\tau = G \times C$ highest-scoring pairs, where $G$ represents the total number of interventions and $C$ denotes the chain size, as detailed in Table~\ref{tab:cluster_statistics_deli_wtd}. For DeliData, we use the mean chain size while for WTD, where dialogues are much longer ($\sim$200) utterances, we use the maximum chain size. Since we only prune pairs for the naive approach, this strategy lets us reduce cross-comparisons at inference while also reducing the impurity of chains by improving recall. 

\section{Further Details on Similarity Baselines}
\label{app:baselines}

\subsection{Token Similarity baseline}
\label{app:tok-sim}
\vspace*{-1mm}

Simple token overlap may indicate correspondence between a probing intervention and its cause(s), as the utterances may share terms. We used the FuzzyWuzzy library \cite{fuzz}, to assess lexical similarity between utterances, using the Levenshtein distance ratio (0--100) between two strings. We computed the token overlap percentage for each probing question and its preceding utterances in both the dev and test sets. Using an empirically-derived threshold from the dev set, based on the average token overlap percentage, if a test pair's token overlap exceeded this threshold, we linked that pair. The computed thresholds for DeliData and WTD were 0.247 and 0.263, respectively.

\subsection{Entity Similarity baseline}
\label{app:ent-sim}
\vspace*{-1mm}

Rather than consider all tokens, which may include semantically irrelevant words, we considered an overlap of salient \textit{entities} between utterances. We computed an Intersection over Union (IoU) of entity counts score based on categorical features derived from task-relevant categories referenced in each utterance. For the DeliData, these categories included vowels, consonants, even digits, and odd digits. For the Weights Task Data, these categories included the five block colors, and their weights, as in \citet{venkatesha2024propositional}. Analogously to the token similarity baseline, we calculated the average IoU between probing interventions and their causal counterparts in the dev set. If a test pair's entity overlap exceeded this threshold, we linked that pair. The computed thresholds for DeliData and WTD were 0.287 and 0.173, respectively.

\subsection{Cosine Similarity Baseline}
\label{app:cos-sim}
\vspace*{-1mm}

Our final non-trained baseline leveraged BERT ({\tt base-uncased}) to generate contextualized sentence embeddings for probing interventions and candidate causal counterparts. This followed the intuition that linked utterances should share some semantic similarity beyond the token or entity level. As with the previous two baselines, we calculated an empirical threshold (average cosine similarity between pairs) from the dev set, being the average of all cosine similarities. If a test pair's cosine similarity exceeded this threshold, we linked that pair. The computed thresholds for DeliData and WTD were 0.597 and 0.644, respectively.




\begin{figure*}
    \centering
    \includegraphics[width=\textwidth]{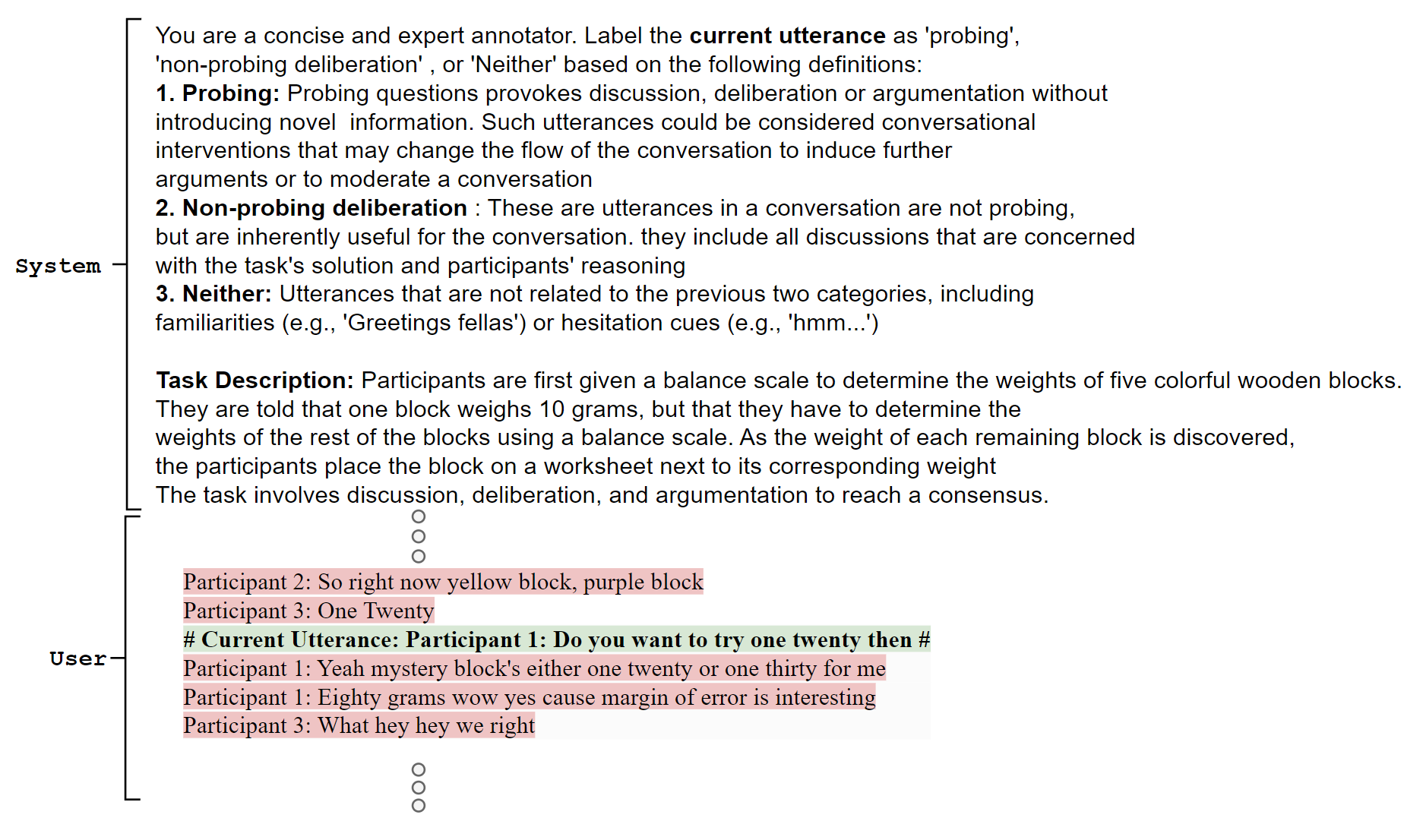}
\vspace*{-2mm}
    \caption{Prompting framework for labeling WTD utterances as probing or non-probing deliberation with GPT.}
    \label{fig:wtd-prompting}
\vspace*{-2mm}
\end{figure*}

\section{Further Details on Label Generation with GPT}
\label{app:label-gen}
\vspace*{-2mm}

Since the Weights Task Dataset does not contain annotations for probing utterances like DeliData does, we also used GPT to label these. Fig.~\ref{fig:wtd-prompting} gives the prompting framework used for this. The probing annotations were also validated by humans (see Appendix~\ref{app:human-eval}).

Algorithm~\ref{alg:gpt_gold_algorithm} provides the iterative labeling algorithm that is used during GPT label generation to assign newly-tagged interventions to the correct preceding cluster.

\begin{algorithm*}[t]
\caption{Gold Cluster Mapping via GPT}\label{alg:gpt_gold_algorithm}
\begin{algorithmic}[1]
\Require $\mathcal{D}$: Sequence of dialogues, $P$: Probing interventions, $GPT(\cdot)$: LLM Prompting Operator 
\State $G \gets \{\}, R \gets []$ \Comment{Gold labels and all GPT responses}

\For{$i = 1$ to $|P|$}
    \State $ctx \gets D[0:P[i].index())$ \Comment{Prior context until probing index}
    \State $r \gets GPT(P[i], ctx)$ \Comment{Generate responses}
    \State $R$.append($r$)
    \If{$i = 1$}
        \For{$resp$ in $r$}
            \State $G[resp] \gets \text{new unique label}$
        \EndFor
        \State $G[P[i]] \gets G[r[0]]$
    \Else
        \State $found \gets \text{False}$
        \For{$resp$ in $r$}
            \If{$resp$ in $G$}
                \State $G[P[i]] \gets G[resp]$
                \State $found \gets \text{True}$
                \State \textbf{break}
            \EndIf
        \EndFor
        \If{not $found$}
        
            \If{$r$ contains element from $P[1:i-1]$}
                \State $idx \gets \text{index of match in } P$
                \State $G[P[i]] \gets G[P[idx]]$
                \For{$resp$ in $r$}
                    \State $G[resp] \gets G[P[idx]]$
                \EndFor
            \Else
                \For{$resp$ in $r$}
                    \State $G[resp] \gets \text{new unique label}$
                \EndFor
                \State $G[P[i]] \gets G[r[0]]$
            \EndIf
        \EndIf
    \EndIf
\EndFor

\State \Return $G$
\Statex Our Gold Cluster Mapping Algorithm iteratively prompts an LLM ($GPT$) to extract causal interventions and rationales. Note that we do not show the rationales generated for each iteration of the loop for space reasons. These generated intervention clusters along with the rationales are then further validated with an exhaustive human evaluation component (see Appendix~\ref{app:human-eval}). 
\end{algorithmic}
\end{algorithm*}

\section{Zero-Shot Prompt Design and Details}
\label{app:zero-shot}
\vspace*{-2mm}

For zero-shot evaluation with LLaMA 2-7B-chat, we designed three prompts, depending on if the paired sentences were gold-labeled as causal and probing, causal and causal, or probing and probing. These prompts were designed to lead the model to a better chance at the correct conclusion, given that the relation between a causal and probing intervention is qualitatively different from that of two causals to a probing intervention that occurs elsewhere, or two probing interventions that share a cause. The prompts are given below.

\begin{tcolorbox}[title=\textsc{\small LLaMA-2-7B-chat Zero-shot Prompt Format: Causal-Probing}]
\begin{dialogue}
	\speak{\sc system\_prompt:} Think step by step. You need to identify if one utterance in a dialogue is going to cause the other utterance to emerge later in the dialogue. Answer in one word: yes or no.
			 
	\speak{\sc user\_prompt:} sentence\_1: \{sentence\_1\}\\ sentence\_2: \{sentence\_2\} 
\end{dialogue}
\end{tcolorbox}

\begin{tcolorbox}[title=\textsc{\small LLaMA-2-7B-chat Zero-shot Prompt Format: Causal-Causal}]
\begin{dialogue}
	\speak{\sc system\_prompt:} Think step by step. You need to identify if these two utterances in a dialogue are going to cause a probing question to emerge later in the dialogue. Answer in one word: yes or no.
			 
	\speak{\sc user\_prompt:} sentence\_1: \{sentence\_1\}\\ sentence\_2: \{sentence\_2\} 
\end{dialogue}
\end{tcolorbox}

\begin{tcolorbox}[title=\textsc{\small LLaMA-2-7B-chat Zero-shot Prompt Format: Probing-Probing}]
\begin{dialogue}
	\speak{\sc system\_prompt:} Think step by step. You need to identify if these two utterances in a dialogue have been caused to emerge by the same preceding utterance in the dialogue. Answer in one word: yes or no.
			 
	\speak{\sc user\_prompt:} sentence\_1: \{sentence\_1\}\\ sentence\_2: \{sentence\_2\} 
\end{dialogue}
\end{tcolorbox}

For a small number of samples (DeliData: 21 out of 7,079, or $\sim$0.297\%; WTD: 232 out of 10,761, or $\sim$2.156\%), LLaMA 2 would not directly provide an answer to the question before reaching the maximum generation length. These were discarded from evaluation.

Due to profanity in a single utterance in the WTD (``{\it So ten plus ten is twenty, twenty plus ten is thirty, thirty plus twenty is fifty, so mystery block's eighty, so I was fucking right}''), LLaMA 2, which is known for its guardrails, would not process 7 pairs  (out of 7,079 $\approx$ 0.099\%) containing this utterance, citing offensive language or ethical or moral standards. These samples were discarded. The limitations inherent in evaluating LLMs on such a PG-13 dataset should be noted.

\section{Additional Results Tables}
\label{app:addl-results}

In coreference tasks, choice of metric bears heavily on the results. Tables~\ref{tab:deli_full_results_test} and \ref{tab:wtd_full_results_test} present results on our two test sets according to the MUC, $B^3$, $CEAF_e$, and CoNLL $F_1$ cluster metrics. 

\begin{table*}[!ht]
    \centering
     
 \footnotesize
    \resizebox{\textwidth}{!}{
    \begin{tabular}{@{}lllrrrrrrrrrrrrrrrrr@{}}
    \toprule
    &&& \multicolumn{3}{c}{MUC} && \multicolumn{3}{@{}c@{}}{$B^3$} & & \multicolumn{3}{c}{$CEAFe$} && CoNLL\\
    \cmidrule{4-6} \cmidrule{8-10} \cmidrule{12-14} \cmidrule{16-16}
    &&& R & P & $F_1$ && R & P & $F_1$ && R &P & $F_1$ && \multicolumn{1}{r}{$F_1$}  \\ 
   \midrule

        

       & Lexical Overlap && 18.2  & 56.4 & 27.5 && 26.6 & 81.3 & 40.0 && 43.9 & 11.6 & 18.3 && 28.6 \\\
       
       & Entity Overlap && 38.6 & 64.0 & 48.2 && 34.9 &  71.7 & 46.9 && 48.8 & 18.5  & 26.8 &&  40.6 \\\
 
      & BERT-Cosine  && 98.9  & 88.4 & 93.4 && 98.6 & 49.9 & 66.3 && 36.0 & 71.3 & 47.8 && 69.2 \\
     & LongContext && 85.9  & 87.5 & 86.7 && 84.7 & 60.7 & 70.7 && 48.9 & 45.6 & 47.2 && 68.2 \\
      & Bidirectional  && 90.7  & 88.3 & 89.5 && 90.8 & 59.2 & 71.7 && 48.6 & 54.7 & 51.4 && 70.9 \\
       &LLaMA 2-7B-chat  && \textbf{99.9}  & 88.5 & \textbf{93.8} && \textbf{99.9} & 49.7 & 66.4 && 35.9 & \textbf{77.1} & 49.0 && 69.7 \\
        \cdashline{1-20}
       & --- Ours (Joint - $W$)  && 92.7  & 89.2 & 90.9 && 92.3 & 60.5 & 73.1 && 52.1 & 62.4 & 56.8 && 73.6  \\
          & --- Ours (Joint + $W$)  && 88.1  & \textbf{91.5} & 89.8 && 87.8 & \textbf{72.6} & \textbf{79.5} && \textbf{64.4} & 55.9 & \textbf{59.9} && \textbf{76.4} \\


    \bottomrule
    \end{tabular}}
    \caption{DeliData test set results. ``LongContext'' denotes \citet{caciularu2021cdlm}'s coreference methodology applied to deliberation chain clustering. ``Bidirectional'' denotes \citet{ahmed20232}'s methodology.}
    \label{tab:deli_full_results_test}
\end{table*}

\begin{table*}[!ht]
    \centering
     
 \footnotesize
    \resizebox{\textwidth}{!}{
    \begin{tabular}{@{}lllrrrrrrrrrrrrrrrrr@{}}
    \toprule
    &&& \multicolumn{3}{c}{MUC} && \multicolumn{3}{@{}c@{}}{$B^3$} & & \multicolumn{3}{c}{$CEAFe$} && CoNLL\\
    \cmidrule{4-6} \cmidrule{8-10} \cmidrule{12-14} \cmidrule{16-16}
    &&& R & P & $F_1$ && R & P & $F_1$ && R &P & $F_1$ && \multicolumn{1}{r}{$F_1$}  \\ 
   \midrule

        

       & Lexical Overlap && 38.6  & 55.1 & 45.4 && 41.6 & 50.0 & 45.4 && 30.3 & 13.8 & 18.9 && 36.6 \\\
       
       & Entity Overlap && 17.1 & 42.2 & 24.4 && 27.2 &  70.0 & 39.2 && 36.1 & 10.7  & 16.5 &&  26.7 \\\
 
      & BERT-Cosine  && \textbf{100.0} & 80.9 & \textbf{89.5} && \textbf{100.0} & 7.1 & 13.2 && 1.7 & \textbf{30.3} & 3.3 && 35.3 \\
      
     & LongContext&& 76.4 & 74.8 & 75.6 && 72.1 & 23.8 & 35.8 && 24.0 & 26.2 & 25.0 && 45.5 \\

      & Bidirectional && 65.7 & 73.0 & 69.2 && 64.5 & 31.5 & 42.4 && 25.4 & 18.2 & 21.2 && 44.3 \\
    &LLaMA 2-7B-chat  && \textbf{100.0}  & 80.9 & \textbf{89.5} && \textbf{100.0} & 7.1 & 13.2 && 1.7 & \textbf{30.3} & 3.3 && 35.3  \\
      \cdashline{1-20}
       & --- Ours (Joint - $W$)  && 50.0 & \textbf{83.3} & 62.5 && 54.4 & \textbf{75.0} & 63.0 && 45.6 & 17.5 & 25.3 && 50.3 \\
          & --- Ours (Joint + $W$)  && 67.9 &	81.9 & 74.2 &&	67.9 &	61.7&	\textbf{64.7} && \textbf{47.5} &	28.2 &	\textbf{35.4} &&	\textbf{58.1} \\



    \bottomrule
    \end{tabular}}
    \caption{WTD test set results with all methods. ``LongContext'' denotes \citet{caciularu2021cdlm}'s coreference methodology applied to deliberation chain clustering. ``Bidirectional'' denotes \citet{ahmed20232}'s methodology..}
    \label{tab:wtd_full_results_test}
\end{table*}

Our method performs well on all metrics, including restrictive ones like $CEAF_e$. We underperform some competing baselines on MUC, but this can largely be attributed to the permissiveness of the MUC metric. We observe that, given the threshold mechanism for the BERT-Cosine baseline, $\sim$40\% of pairs in both test sets are labeled as positives by default.  Given that the resulting false positives link to interventions that have true links to a larger chain, the transitive closure mechanism tends to link most or all utterances into a single intervention cluster. This is reflected in the 100\% or near-100\% recall achieved by BERT-Cosine and LLaMA 2 zero-shot in both MUC {\it and} $B^3$, and the extremely low $CEAF_e$ recall due to $CEAF_e$'s assumption that each key entity should only be mapped to a single reference entity. This indicates that while MUC especially is foundational in coreference, it may be a less useful metric in deliberation chain construction.

We currently exclude the LEA metric from our evaluation metrics for two reasons. First, we use gold intervention labels since the current work only considers link assignment to pairs in building deliberation chains and \textit{not} intervention detection. Moreover, assigning an ``importance'' measure to various interventions at a linguistic level is beyond the scope of the paper. As such, for a fair evaluation between commonly used metrics, we focus on CoNLL $F_1$ as the average of the MUC, $B^3$ and $CEAF_e$ $F_1$ scores. By contrast, LEA specifically re-weighs evaluations to mitigate the ``mention identification effect'' and would apply a task-irrelevant importance measure to interventions \cite{moosavi2016coreference}. We leave determination of optimal metrics for this task to future work.

\section{Human Evaluation of GPT-Annotated Labels}
\label{app:human-eval}

\begin{figure}[h!]
    \centering
    \includegraphics[width=.45\textwidth]{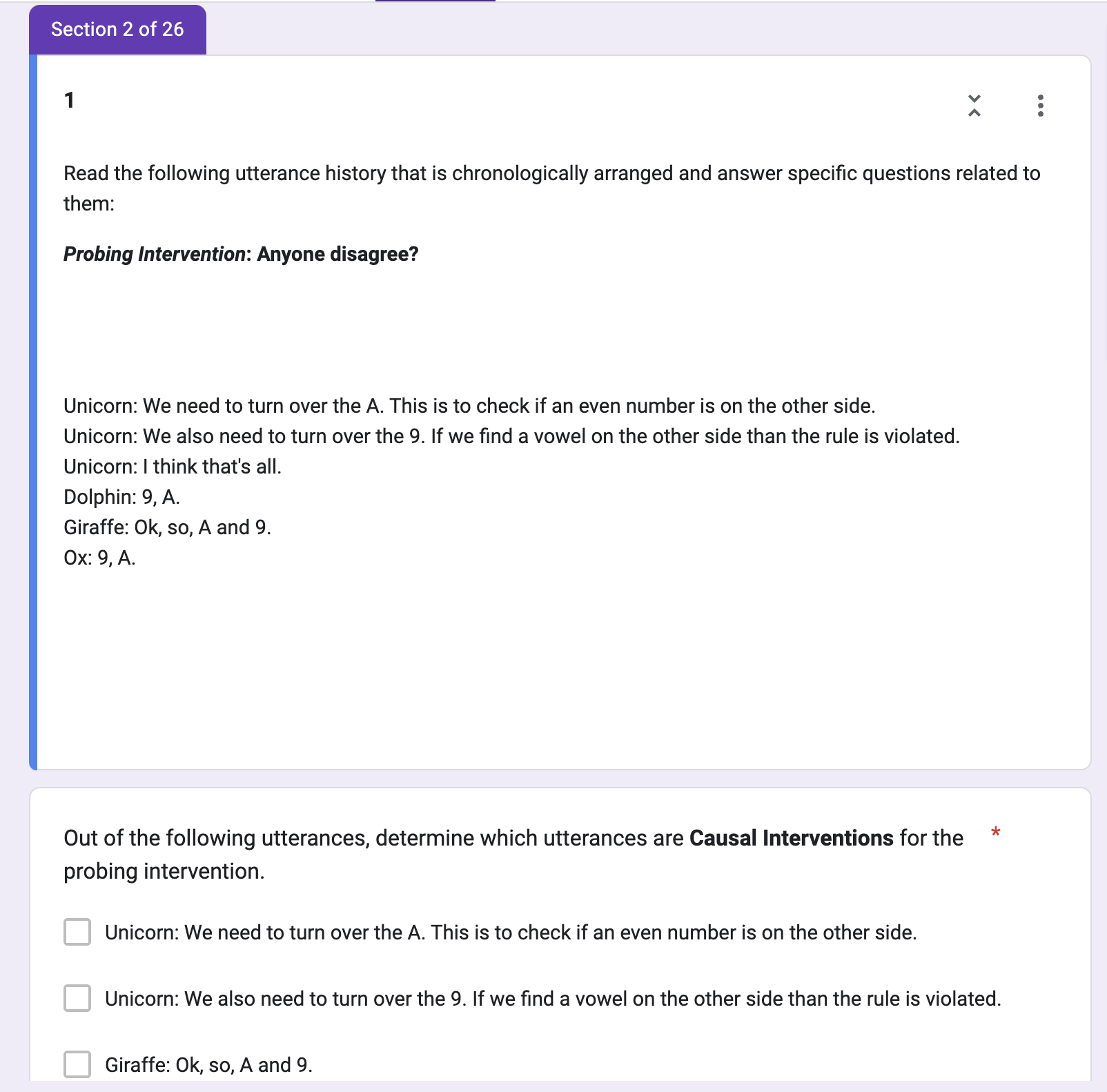}
    \caption{Causal intervention sample presented to evaluators.}
    \label{fig:rationale-eval}
\end{figure}

Four evaluators (all adult English speakers) took a survey containing probing interventions and candidate causal interventions (25 sets each drawn from the WTD and DeliData corpora), the ground truth label (which was also given to GPT 3.5-turbo for generation), and the generated inner monologue FTR (see Fig.~\ref{fig:rationale-eval}). They were asked to answer seven multiple choice questions for each sample, designed to explore various aspects of the dialogue explanation.
 
The questions included:

\begin{itemize}
    \item {\bf Relevance to Context}: Are the Causal Intervention(s) relevant to the context? ({\it Yes}/{\it No})
    \item {\bf Presence in Sequences}: Are the Causal Interventions(s) present in the sequences of utterances? ({\it yes}/{\it no}/{\it not enough information})
    \item {\bf Information Sufficiency}: How much information do the Causal Interventions(s) have, to justify them being actual causal interventions? ({\it enough}/{\it not enough}/{\it more than enough}/{\it can't say})
    \item {\bf Acceptability/Plausibility}: Are the Causal Interventions acceptable or plausible considering the context? ({\it yes}/{\it no}/{\it can't say})
    \item {\bf Overlap with Own Explanations}: If you were to use your own explanations for selecting the causal interventions, how much of an overlap does your thought-pattern have with the given rationales? ({\it high overlap}/{\it some overlap}/{\it minimal overlap}/{\it no overlap})
\end{itemize}
The statistics for the chain lengths of the drawn samples are as follows:


\begin{table}[htb]
    \centering
     
    \resizebox{\linewidth}{!}{
    \begin{tabular}{@{}llcc@{}}
        \toprule
        & & \multicolumn{2}{c}{Chain Length Statistics} \\
        \cmidrule(lr){3-4}
        & & DeliData & WTD \\
        \midrule
        
        Min Chain Length & & 3 & 3 \\
        Max Chain Length & & 8 & 10 \\
        Mean Chain Length & & 5.65 & 5.3 \\
        
        \bottomrule
    \end{tabular}
    }
    \caption{Chain length statistics of the human evaluation samples.}
\end{table}

These chain lengths indicate that the sampled probing interventions are representative of the respective test sets, as their mean chain lengths align with the dataset averages, and the distributions are within the expected ranges.

Annotations were performed by members of the authors' research lab in the course of their normal duties. A different pair of annotators was used to assess samples from each corpus. Three annotators were male and one, female. Annotators had no prior experience in the task. The survey was determined to be Not Human Subjects Research by the institutional review board.



The survey response results shown in Fig.~\ref{fig:data-wtd-deli-eval} show that the causal interventions, for both the WTD and DeliData, have positive valences when evaluated for relevance to the context, presence in sequence, and acceptability/plausibility. These positive scores show high association between causal interventions and their respective probing questions, given the context of the utterance history. DeliData information sufficiency is rated higher than WTD’s, which shows that the DeliData contained more information to support the justification of classifying utterances as causal interventions. This could be a reflection of the use of different cards for the Wason Selection Task between groups in the DeliData experiment; where the WTD experiments utilized the same task items across all experiments, resulting in more repetitive phrases.

\end{document}